\documentclass[letterpaper]{article} 
\usepackage{aaai25}  
\usepackage{times}  
\usepackage{helvet}  
\usepackage{courier}  
\usepackage[hyphens]{url}  
\usepackage{graphicx} 
\usepackage{natbib}  
\usepackage{caption} 
\usepackage{algorithm}
\usepackage{algorithmic}
\usepackage{xcolor}

\usepackage{modroman}
\usepackage{subfigure}
\usepackage{amsmath}
\usepackage{amsfonts}
\usepackage{multirow}
\usepackage{caption}
\usepackage{subcaption}
\usepackage{booktabs}

\newtheorem{definition}{Definition}

\usepackage{newfloat}
\usepackage{listings}
\DeclareCaptionStyle{ruled}{labelfont=normalfont,labelsep=colon,strut=off} 
\floatstyle{ruled}
\newfloat{listing}{tb}{lst}{}
\floatname{listing}{Listing}
\title{Normalize Then Propagate: Efficient Homophilous Regularization for Few-shot Semi-Supervised Node Classification}
\author{
    Baoming Zhang\textsuperscript{\rm 1},
    Mingcai Chen\textsuperscript{\rm 2},
    Jianqing Song\textsuperscript{\rm 1},
    Shuangjie Li\textsuperscript{\rm 1},
    Jie Zhang\textsuperscript{\rm 1},
    Chongjun Wang\textsuperscript{\rm 1}\footnote{Corresponding author.}
}
\affiliations{
    \textsuperscript{\rm 1} State Key Laboratory of Novel Software Technology, Nanjing University, China    \\
    \textsuperscript{\rm 2} Nanjing University of Posts and Telecommunications, China                       \\

    zhangbm@smail.nju.edu.cn, chenmc@njupt.edu.cn, \{sjq, shuangjieli\}@smail.nju.edu.cn, guyuexiao95@gmail.com, chjwang@nju.edu.cn
}

\usepackage{bibentry}

\begin{document}

\maketitle

\begin{abstract}
    Graph Neural Networks (GNNs) have demonstrated remarkable ability in semi-supervised node classification.
    However, most existing GNNs rely heavily on a large amount of labeled data for training, which is labor-intensive and requires extensive domain knowledge.
    In this paper, we first analyze the restrictions of GNNs generalization from the perspective of supervision signals in the context of few-shot semi-supervised node classification.
    To address these challenges, we propose a novel algorithm named NormProp, which utilizes the homophily assumption of unlabeled nodes to generate additional supervision signals, thereby enhancing the generalization against label scarcity.
    The key idea is to efficiently capture both the class information and the consistency of aggregation during message passing, via decoupling the direction and Euclidean norm of node representations.
    Moreover, we conduct a theoretical analysis to determine the upper bound of Euclidean norm, and then propose homophilous regularization to constraint the consistency of unlabeled nodes.
    Extensive experiments demonstrate that NormProp achieve state-of-the-art performance under low-label rate scenarios with low computational complexity.

\end{abstract}

%

\section{Introduction}
Graphs are powerful data structures that allow us to easily express various relationships (i.e., edges) between objects (e.g., nodes).
Graph Neural Networks (GNNs) \cite{nips2016chebnet, DBLP:conf/iclr/2017/gcn, iclr2018gat} are a family of machine learning models specifically designed to model non-Euclidean graph data.
Modern GNNs have emerged as the forefront approach for acquiring representations at various levels within graphs,
encompassing node-level \cite{DBLP:conf/iclr/2017/gcn, iclr2019appnp}, edge-level \cite{nips2018link-GNN}, and graph-level tasks \cite{nips2018diff-pool, icml2022g-mixup}.
A typical GNN model usually follows the message passing framework \cite{icml2017mpnn}, which mainly contains two operators: feature transformation and feature propagation.
In semi-supervised node classification settings \cite{zhu2002label-propagation}, only a subset of nodes have labels, and the parameters of the GNN are trained using the classification loss function that relies on the representations of labeled nodes \citep{DBLP:conf/iclr/2017/gcn}.

However, in numerous real-world scenarios of semi-supervised learning, obtaining a substantial amount of accurate labels for training can be challenging due to the labor-intensive data annotation \cite{nips2019self-train4few-shot}.
Furthermore, in the problem of semi-supervised node classification  with limited labeled nodes \cite{aaai2022meta-pn}, the scarcity of labeled nodes per class not only exacerbates that GNNs are prone to overfitting, but also magnifies additional challenge.
Recent studies \citep{ma2022partition-gnn, nips2020graph-policy, cikm2020degree-bias} have shown that GNNs exhibit certain biases originating from the perspective of graph topology.
\citet{ma2021subgroup-fair} study the subgroup generalization of GNNs and find that the shortest path distance to labeled nodes can also affect the GNNs’ performance.
Specifically, \citet{nips2024label-position-bias} uncover the label position bias, which indicates that \textit{the node closer to the labeled nodes tends to perform better}.
Considering the citation networks Cora and Citeseer (Planetoid split) \cite{icml2016planetoid-dataset}, the label signal is only relevant to $61.4\%$ and $32.8\%$ of the nodes, respectively, in the case of two-layer GNN.
When only a few nodes are labeled, it exacerbates label position bias, as a larger number of nodes lack adequate label information during message passing. 

To mitigate the problem of overfitting due to limited labels, a straightforward idea is to enlarge the receptive field of aggregation by incorporating additional representations from unlabeled high-order nodes into the calculation of the classification loss.
Unfortunately, most of the existing GNNs have been designed with shallow architectures primarily due to the \textit{over-smoothing issue} \cite{nips2021dirichlet-gnn}.
With the increasing of GNN layers, the Dirichlet energy converge to zero \cite{icml2020over-smoothing}. As a result, node representations become increasingly similar, blurring the distinction between different classes.
Moreover, high-order neighbor aggregation further increases the complexity of message passing and may fail to acquire information from remote nodes on another connected component.

\begin{figure}[t]
    \centering
    \includegraphics[width=0.95\linewidth]{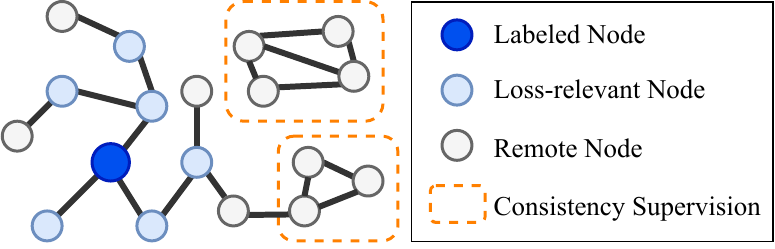}
    \caption{The source of the loss signal for a two-layer GNN. Suppose that there is only one labeled node in the graph, and that the graph consists of two connected components.}
    \label{fig:intro}
\end{figure}

An example on a two-layer GNN \cite{DBLP:conf/iclr/2017/gcn} is shown in Figure \ref{fig:intro}.
The receptive field of classification loss solely focuses on the labeled node and its two-hop neighbors (loss-relevant nodes).
We refer to nodes that do not participate in loss calculation as \textit{remote nodes}.
Previous works on limited labels in GNN mainly utilize remote nodes from three perspectives: high-order neighbors aggregation \cite{aaai2022meta-pn}, pseudo labeling \cite{aaai2020M3S} and graph structure learning \cite{ijcai2023violin}.
However, whether using pseudo-label iterative training or establishing connections between labeled nodes and remote nodes, there is a higher computational cost.
In order to acquire more training signals for GNN with low complexity, we directly generate supervision signals from remote nodes instead of using message-passing framework to aggregate more information from unlabeled nodes.
As shown in Figure \ref{fig:intro}, additional supervision signals are derived from the neighbor consistency of remote nodes, based on the classic homophily assumption \cite{arxiv2019low-pass-filter}.

To address the above challenges, we propose \textbf{NormProp}, a novel GNN model that enhances the model's generalization ability by incorporating the consistency information from commonly ignored nodes.
With the aim of efficiently obtaining a consistency metric synchronized with message passing, our key idea is to \textit{decouple the direction and norm of node representation vectors}.
The direction and Euclidean norm of node representation vector, extracted by NormProp, respectively denote the class information and neighbor consistency.
Moreover, the core operation is ``\textbf{Norm}alize then \textbf{Prop}agate'', which maps node features to the unit hypersphere and subsequently applies a low-pass filter \cite{icml2019sgc}.
The classification loss relies on computing the cosine similarities between the labeled node representations and their corresponding class prototypes.
Additionally, We conduct a theoretical analysis to determine the lower and upper bounds of Euclidean norm for representation vectors after propagation.
Through the above theory, we propose \textbf{Homophilous Regularization} based on homophily assumption, which constrains the neighbor consistency of remote nodes.
Training supervision involves both the classification loss incurred from labeled nodes and the homophilous regularization applied to the unlabeled nodes.
Extensive experiments on node classification tasks with limited labeled nodes demonstrate that NormProp not only exhibits excellent performance but also requires low computational complexity.

In summary, we list our contributions as follows:
\begin{itemize}
    \item We analyze the problem of few-shot semi-supervised node classification from the perspective of supervision signals, and consequently provide an efficient idea to enhance the generalization by leveraging the homophily of remote nodes.
    \item We propose a novel GNN, NormProp, where the direction and Euclidean norm of node representations learned by our model respectively indicate class information and consistency of aggregation.
    We conduct theoretical analysis on the upper bound of Euclidean norm, and propose homophilous regularization to constrain the neighbor consistency of unlabeled nodes.
    \item Extensive experimental results indicate the effectiveness and efficiency of NormProp, providing compelling validation for the advantages of NormProp over state-of-the-art methods.
\end{itemize}

\section{Preliminaries and Related Works}
\subsection{Notations}
We consider transductive semi-supervised unweighted and undirected graphs node classification.
Given a graph $G = (V, E)$ with $n = |V|$ nodes and $m = |E|$ edges, where $V = \{v_1, v_2, \cdots, v_n\}$ is the set of nodes and $E \subseteq V \times V$ is the set of edges between nodes in $V$.
Let $X \in \mathbb{R}^{n \times d}$ denote the node feature matrix, with $x_i$ presenting a $d$-dimensional feature vector of node $v_i$.
The topology information of the entire graph is described by the adjacency matrix $\mathbf{A} \in \mathbb{R}^{n \times n}$, where $\mathbf{A}_{ij} = 1$ if $\left<v_i, v_j\right> \in E$, otherwise $\mathbf{A}_{ij} = 0$. Let $\mathcal{N}(i)$ denote the neighboring nodes set of node $v_i$.
The diagonal degree matrix is denoted as $\mathbf{D} \in \mathbb{R}^{n \times n}$, where $D_{ii} = |\mathcal{N}(i)|$ represents the degree of node $v_i$.
Moreover, the adjacency matrix and degree matrix of graph with self-loops are respectively denoted as $\tilde{\mathbf{A}} = \mathbf{A} + \mathbf{I}_n$ and $\tilde{\mathbf{D}} = \mathbf{D} + \mathbf{I}_n$.
For few-shot semi-supervised node classification, only a few training nodes have labels.

\subsection{Decoupled GCN}
\label{sec:gnn}
GCN \cite{DBLP:conf/iclr/2017/gcn} stacks layers of first-order spectral filters, with each message-passing layer followed by a learnable linear transform $\mathbf{W}^{(k)}$ and a nonlinear activation function $\sigma(\cdot)$.
The $k$-th layer of GCN is defined as follow:
\begin{equation}
\label{eq:gcn}
    \mathcal{H}^{(k + 1)} = \sigma \left( \hat{\mathbf{A}} \mathcal{H}^{(k)} \mathbf{W}^{(k)} \right)  \;,
\end{equation}
where $\hat{\mathbf{A}} = \tilde{\mathbf{D}}^{-\frac{1}{2}} \tilde{\mathbf{A}} \tilde{\mathbf{D}}^{-\frac{1}{2}}$ is the symmetrically normalized adjacency matrix,
$\mathcal{H}^{(k)}$ denotes node representations of the $k$-th GCN layer.
Recent decoupled GCN separate the transformation and propagation in each layer, such as APPNP \cite{iclr2019appnp} and GPR-GNN \cite{chien2020gpr-gnn}.
Formally, the decoupled GCN can be written as:
\begin{equation}
\label{eq:decoupled-gcn}
    \hat{Y} = \mathrm{softmax} \left( \bar{\mathbf{A}} f_\theta \left( X \right) \right) \;,
\end{equation}
where $f_\theta (\cdot)$ is a feature encoder, $\bar{\mathbf{A}} = \sum \alpha_k \hat{\mathbf{A}}^k$ is the propagation strategy, which involves the $k$-order neighbors.
Our theoretical analysis  is based on the decoupled GCN form.

\begin{figure*}[t]
    \centering
    \includegraphics[width=0.85\linewidth]{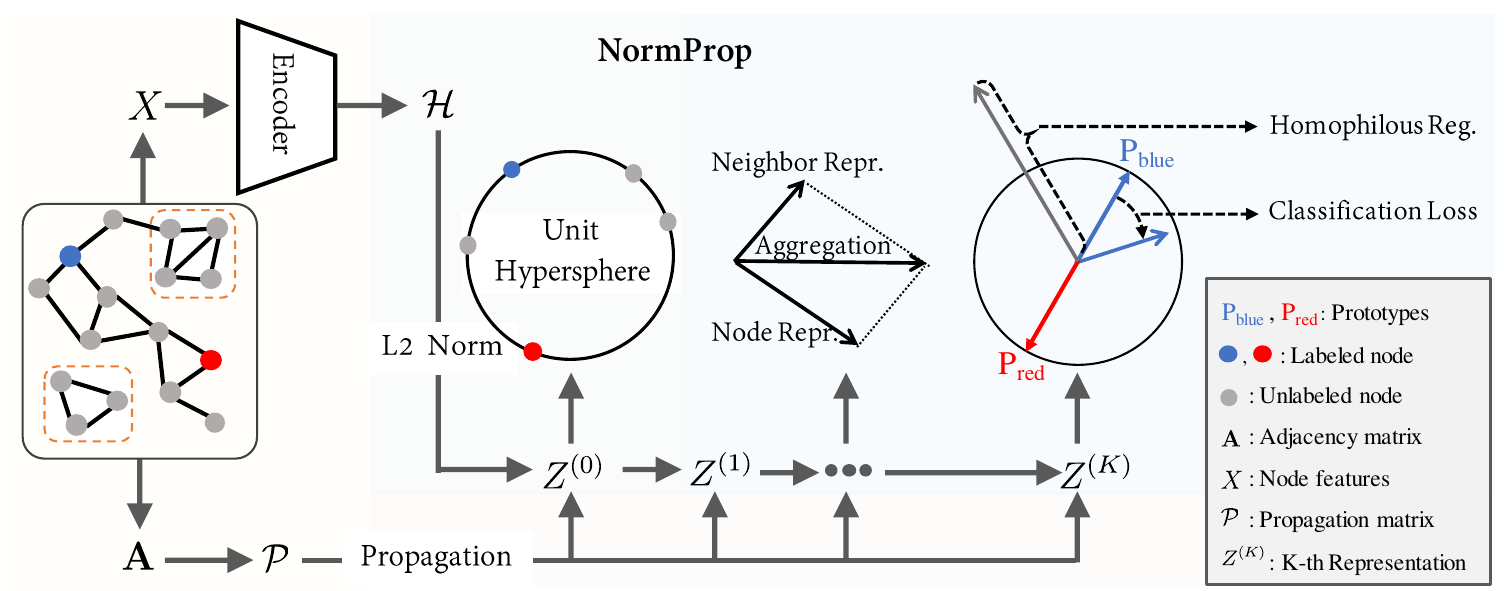}
    \caption{An illustration of the proposed NormProp (Normalize then Propagate).}
    \label{fig:hyperprop}
\end{figure*}

\subsection{Related Works}
As mentioned in the introduction, with only a few labeled nodes, the expressiveness of GNNs to utilize information in graph data is limited.
M3S \cite{aaai2020M3S} propose a multi-stage training framework, adding confident data with virtual labels to the labeled set to enlarge the training set.
Meta-PN \cite{aaai2022meta-pn} infers high-quality pseudo labels on unlabeled nodes with a meta-learned label propagation strategy, augmenting rare labeled nodes and enabling broad receptive fields during training.
Currently, there are some techniques available to modify structure and enlarge receptive field, thereby amplifying the influence of label signals.
To counter the over-smoothing issue, decoupled GCNs have gained popularity due to their simplicity.
APPNP \cite{iclr2019appnp} combines GNN with the concept of Personalized PageRank \cite{Page1999pagerank}, effectively decoupling the feature transformation and propagation with restart probability.
DAGNN \cite{kdd2020dagnn} aggregate all the previous embeddings at the final layers.
The most time-consuming part of many GNNs is propagation \cite{chen2018fastgcn}, and expanding the scope of propagation will aggravate the problem.
Graph Structure Learning (GSL) \cite{zhu2021graph-structure-learning} can additionally facilitate the propagation of label signals.
AM-GCN \cite{kdd2020am-gcn} propose an adaptive multi-channel GCN that combines the original structures with the $k$NN graph generated based on the similarity of node features.
Violin \cite{ijcai2023violin} creates virtual connections between labeled and unlabeled nodes.
However, many GSL methods require adjusting graph structure by modeling nodes similarity, which entails quadratic complexity.

\section{NormProp: Normalize then Propagate}

We utilize the decoupled GCN framework, which separates representation transformation and propagation (Eq.\ref{eq:decoupled-gcn}).
The above-mentioned framework eliminates the linear transformation and non-linear activation in each graph convolution layer, enabling us delve deeper into the node representation changes caused by propagation, especially consistency of the nodes encountered during propagation.

Expanding on the aforementioned decoupled GCN framework, we map node features to embeddings in the hyperspherical space, and present a new approach called NormProp.
In our approach, we propagate $\mathrm{L}_2$ hypersphere embeddings and generate supervision signals from both directions (class information) and Euclidean norm (consistency of aggregation), as illustrated in Fig.\ref{fig:hyperprop}.
Unlike many few-shot node classification methods that rely on pseudo-labels \cite{aaai2022meta-pn} and multi-stage training \cite{aaai2020M3S}, the proposed NormProp is end-to-end.
Moreover, the proposed framework consists of the following parts:

(\,\nbRoman{1}\,) \textbf{Hyperspherical Node Feature Encoder.}
We map the vanilla node feature onto a unit hypersphere.
Precisely, we apply a multilayer perceptron $f_\phi : \mathbb{R}^d \rightarrow \mathbb{R}^{d'} $ operates only on nodes' feature, and then generate $\mathcal{H} = [ \mathcal{H}_1^\mathrm{T}, \cdots, \mathcal{H}_n^\mathrm{T} ] \in \mathbb{R}^{n \times d'}$.
Subsequently, we project network outputs $\mathcal{H}$ to the hypersphere through $\mathrm{L}_2$ normalization, resulting the normalized node representations $Z^{(0)} \in \mathbb{R}^{n \times d'}$.

(\,\nbRoman{2}\,) \textbf{Propagation.}
Since we apply the decoupled GCN framework in order to better study the impact of propagation, the propagation corresponds to a fixed low-pass filter \cite{arxiv2019low-pass-filter} applied to each node hyperspherical representation dimension, akin to the approach employed in SGC \cite{icml2019sgc}.
The ultimate representation of node $v_i$ is essentially the weighted sum of its $K$-order neighborhoods' hyperspherical representation, e.g. , $Z^{(K)}_i = \sum_{j \in \mathcal{N}^K(i)} \, \alpha_{ij} Z^{(0)}_j$, where $\mathcal{N}^K(i)$ denotes the $K$-order neighbors of node $v_i$.

In summary, the final representation of the nodes extracted through our NormProp can be succinctly expressed using the following mathematical formula:
\begin{flalign}
\label{eq:encode}
&& &\mathcal{H}_i = f_\phi \left( X_i \right) &&\rhd \text{Encode node features \;}
\\
\label{eq:l2-norm}
&& &Z^{(0)}_i = \frac{\mathcal{H}_i}{\big\Vert \mathcal{H}_i \big\Vert_2} &&\rhd \text{$\mathrm{L}_2$ Normalization}
\\
\label{eq:propagate}
&& &Z^{(K)} = \mathcal{P}^K Z^{(0)} &&\rhd \text{Propagation}
\end{flalign}
where $X_i \in \mathbb{R}^{d}$ denotes the vanilla feature of node $v_i$,
$\mathcal{H}_i \in \mathbb{R}^{d'}$ denotes the non-normalized representation of $v_i$ obtained by MLP $f_\phi(\cdot)$ with parameter set $\phi$,
$Z^{(0)} \in \mathbb{R}^{n \times d'}$ denote the hyperspherical representations after $\mathrm{L}_2$ normalization,
$Z^{(K)} \in \mathbb{R}^{n \times d'}$ denotes the final node representations achieved by aggregating $K$-order neighbors, and the corresponding propagation weight matrix is denote as $\mathcal{P}^K \in \mathbb{R}^{n \times n}$.
As shown in Fig.\ref{fig:hyperprop}, the representation vector $Z^{(K)}$ can be decouple into direction and Euclidean norm, corresponding to class information and consistency of aggregation, respectively.
Similar to Eq.21 in Appendix A.1, classification is performed based on the final node representation $Z^{(K)}$ by finding the nearest class prototype in $\mathbb{P} = \left[\mathrm{P}_1, \cdots, \mathrm{P}_C \right] \in \mathbb{R}^{C \times d'}$, using \textit{cosine similarity} as the metric. Formally, the inference of node $v_i$ can be defined as:
\begin{equation}
\label{eq:proto-cos}
    c^* = \operatorname*{arg\,max}_{c \, \in \, C} \left( \mathop{cos} \theta_{(Z^{(K)}_i \, , \, \mathrm{P}_c)} \right) \, .
\end{equation}
Additionally, we implement the predefined hyperspherical prototypes following Mettes et al. \cite{nips2019hpn}. The solution of the
hyperspherical prototypes is achieved through efficient data-independent optimization, as briefly described in Appendix A.1.

\section{Homophilous Regularization: Consistent Loss based on Euclidean Norm}

In this section, we establish the upper bound of Euclidean Norm in our NormProp, which is equivalent to the random walk on the graph.
In the following, in order to further enhance the consistency of certain subgraphs, we propose the homophilous regularization based on homophily assumption.
Finally, we take a different angle and analyze that homophilous regularization fundamentally symbolizes the agreement between original graph and augmented graph across propagation.

\subsection{Upper Bound of Euclidean Norm}
\label{sec:norm-upper-bound}

In the previously proposed framework, the directions of node representations $Z^{(K)}$ effectively describe their class information using the metric of cosine similarity.
In other ways, we perform a comprehensive analysis to further explore the relationship between the Euclidean Norm of $Z^{(K)}$ and the consistency of node representations encountered during message passing.

\begin{definition}
[Normed Vector Space]
\label{the:normed-vec-space}

Let $E$ be a vector space over a field $K$, where $K$ is either the field $\mathbb{R}$ of reals, or the field $\mathbb{C}$ of complex numbers. A norm on $E$ is a function $\Vert \cdot \Vert : E \rightarrow \mathbb{R}_{+}$ , assigning a nonnegative real number $\Vert u \Vert$ to any vector $u \in E$, and satisfying the following conditions for all $x, y, z \in E$ :
\begin{flalign}
&& &\Vert x \Vert \geq 0                                            &&\rhd \text{Positivity}
\\
&& &\Vert \lambda x \Vert = \vert \lambda \vert \, \Vert x \Vert    &&\rhd \text{Scaling}
\\
&& &\Vert x + y \Vert \leq \Vert x \Vert + \Vert y \Vert            &&\rhd \text{Triangle inequality\;}
\end{flalign}
A vector space $E$ together with a norm $\Vert \cdot \Vert$ is called a normed vector space.

\end{definition}

In this paper, we consider Euclidean Norm on the real field, specifically the $l_2$ norm of node representations during propagation. 
The outcome of the propagation can be understood as a weighted sum of the representations from neighboring nodes.
According to the properties of Definition \ref{the:normed-vec-space}, We can derive an upper bound on the Euclidean norm of the representation propagation:
\begin{equation}
\label{eq:norm-upper1}
    \bigg\Vert \sum_{j \in \mathcal{N}(i) } \alpha_{ij} \cdot Z_j^{(0)} \;\; \bigg\Vert_2
    \leq 
    \sum_{j \in \mathcal{N}(i) } \alpha_{ij} \big\Vert Z_j^{(0)} \big\Vert_2
    \; ,
\end{equation}
where $\sum_{j \in \mathcal{N}(i) } \alpha_{ij} Z_j^{(0)} $ denotes the propagation result of node $v_i$ ;
$\alpha_{ij} \geq 0$ denotes the aggregation coefficent .
As shown in Eq.\ref{eq:l2-norm} , node representations $Z^{(0)}$ are distributed over the surface of hypersphere, so $\big\Vert Z_j^{(0)} \big\Vert_2 = 1$ .
Finally, the $l_2$ norm upper bound of $Z^{K}$ can be formulated as:
\begin{equation}
\label{eq:norm-upper2}
    \Big\Vert \, Z^{(K)}_i \, \Big\Vert_2
    \leq 
    \left[ \mathcal{P}^K \mathbf{1}_{n \times 1} \right]_{i} \; ,
\end{equation}
where $\mathrm{1}_{n \times 1}$ denotes an all-ones matrix of size $n \times 1$ ;
$\mathcal{P}^K \in \mathbb{R}^{n \times n}$ denotes the weight matrix of aggregation.

It is important to highlight that in Eq.\ref{eq:norm-upper2}, the requirement for equality is that all the representation vectors involved in the aggregation are parallel. Furthermore, the Euclidean norm of the final node representation is positively correlated with consistency of aggregation.
Additionally, the upper bound $\mathcal{P}^K \mathbf{1}_{n \times 1}$ in Eq.\ref{eq:norm-upper2} can be interpreted as the result of a random walk \cite{lovasz1993random-walk-on-graph} on the graph.
Because of Eq.\ref{eq:l2-norm}, all nodes have equal norm (unnormalized probability), and the norm of node representations after propagation (Eq.\ref{eq:propagate}) reflects the importance of nodes in the random walk.

\subsection{Homophilous Regularization}
\label{sec:norm-loss}
Building upon the foundation of the bound,
we will introduce the consistent metric based on Euclidean norm, and explore additional supervisory signals from unlabeled nodes.
\begin{definition}
[Consistent Metric]
    For node $v_i$, the lower and upper bounds of its Euclidean norm in the propagation of NormProp could be relaxed as:
    \begin{equation}
        0 \; \leq \; \Big\Vert \, Z_i^{(K)} \Big\Vert_2 \; \leq \; \left[ \mathcal{P}^K \mathbf{1}_{1 \times n} \right]_i \; .
    \end{equation}
    Moreover, because of the non-negativity property of the norm and its positive correlation with consistency of the subgraph, we normalize it using the upper bound. For node $v_i$, we define consistent metric $\zeta \in [0, 1]$ as follow:
    \begin{equation}
    \label{eq:zeta}
        \zeta =
        \frac{
            \big\Vert \, Z_i^{(K)} \big\Vert_2
        }{
            \left[ \mathcal{P}^K \mathbf{1}_{1 \times n} \right]_i
        } \; .
    \end{equation}
\end{definition}

Many theoretical foundations of GNNs are the homophily assumption \cite{ciotti2016homophily-citation}, which states that nodes tend to connect with others who are ``similar'' to them.
However, previous works \cite{nips2021topology-imbalance, iclr2020measuring-graph-info} have pointed out that the nodes close to the class boundary have the risk of information conflict.
Therefore, we consider that the node $v_i$ exhibits a high level confidence, e.g., cosine similarity between $Z^{(K)}_i$ and any prototype $\mathrm{P} \in \mathbb{P}$ is greater than threshold $\tau$, indicating a high level consistency during message passing.
Let $\Omega_\tau = \{ v_j : \mathop{cos} ( \theta_{\mathrm{P} , Z^{(K)}_j} ) \geq \tau \}$. Formally, the homophilous regularization $\mathcal{L}_h$ is defined based on consistent metric $\zeta$:
\begin{equation}
\label{eq:norm-based-loss}
    \mathcal{L}_h = 1 - \frac{1}{| \Omega_\tau |} \sum_{i \in \Omega_\tau}
            \frac{
                \big\Vert \, Z_i^{(K)} \big\Vert_2
            }{
                \left[ \mathcal{P}^K \mathbf{1}_{1 \times n} \right]_i
            }
\end{equation}
Homophilous regularization aims to enhance the consistency of certain specific subgraphs.
The term "Homophilous Regularization" is used to describe $\mathcal{L}_h$ in Eq.\ref{eq:norm-based-loss} because its mathematical form is similar to $L_2$ regularization $\Vert \cdot \Vert_2$.

\subsection{Reinterpretation of Homophilous Regularization}
In Eq.\ref{eq:norm-based-loss}, we view the homophilous regularization as the ratio of Euclidean norm.
Furthermore, we will provide a more intuitive explanation from a different perspective.
$\mathcal{L}_h$ can be expressed over the entire graph as a  matrix formulation:
\begin{equation}
\label{eq:norm-based-loss1}
    \mathcal{L}_h = 1 - \frac{1}{\big\Vert \mathcal{M} \big\Vert_1} \,
        \mathrm{tr} \, \Big( \,  \frac{
                \left(
                    Z^{(K)^\mathbf{T}} Z^{(K)} \, \mathcal{M}
                \right)^{- \frac{1}{2}}
            }{
                \mathcal{P}^K \, \mathbf{1}_{n \times n}
            }
        \, \Big) \; ,
\end{equation}
where $\mathrm{tr}(\cdot)$ denotes trace of a matrix;
$\mathrm{1}_{n \times n}$ denotes an all-ones matrix of size $n \times n$  ;
$\mathcal{M} = \mathrm{diag} ( \mathrm{M}_1, \cdots, \mathrm{M}_n ) \in \mathbb{R}^{n \times n}$ denotes the diagonal masking matrix, $\mathrm{M}_i = 1$ if $v_i \in \Omega_\tau$, or $\mathrm{M}_i = 0$ otherwise.
The portion of Eq.\ref{eq:norm-based-loss1} that is related to the node representations can be transformed into the following form.
\begin{align}
    Z^{(K)^\mathbf{T}} Z^{(K)} \mathcal{M}
    &= ( \mathcal{P}^K Z^{(0)})^\mathbf{T} (\mathcal{P}^K Z^{(0)} ) \mathcal{M} 
    \\
    &= ( \mathcal{P}^K Z^{(0)})^\mathbf{T} \mathcal{P}^K ( Z^{(0)} \mathcal{M})
    \\
    &= ( \mathcal{P}^K Z^{(0)})^\mathbf{T} (\mathcal{P}^{K} \tilde{Z}^{(0)} )
    \; ,
\end{align}
where $\tilde{Z}^{(0)} = Z^{(0)} \mathcal{M}$ denotes feature masking operation \cite{nips2020graphCL}, e.g., sets the norm of nodes with low confidence to zero.
On one hand, $\mathcal{P}^K Z^{(0)}$ denotes the propagation using all nodes, while on the other hand, $\mathcal{P}^{K} \tilde{Z}^{(0)}$ represents the propagation using only the features from high-confident nodes.
Essentially, $\mathcal{L}_h$ serves as the contrastive loss, ensuring the consistency of the inner product between the augmented graph and the original graph.

\section{Model Implementation and Training}

\subsection{Implementation of NormProp}
Following the graph convolution design of GCN \cite{DBLP:conf/iclr/2017/gcn} and SGC \cite{icml2019sgc}, we employ the graph convolution matrix with the renormalization trick as aggregation operator, e.g. , $\mathcal{P} = \tilde{\mathbf{D}}^{-\frac{1}{2}} \tilde{\mathbf{A}} 
\tilde{\mathbf{D}}^{-\frac{1}{2}}$ .
Due to the removal of linear transformation and activation function in message passing in our framework, the final form of propagation is equivalent to the fixed $K$-order low-pass filter in SGC.
During the computation of node representations (Eq.\ref{eq:encode} $\sim$ Eq.\ref{eq:propagate}) in the implementation of NormProp, the propagation rule (Eq.\ref{eq:propagate}) can be written as:
\begin{equation}
\label{eq:hyperGCN}
    Z^{(K)} = \left( \tilde{\mathbf{D}}^{-\frac{1}{2}} \tilde{\mathbf{A}} \tilde{\mathbf{D}}^{-\frac{1}{2}} \right)^K Z^{(0)} \, ,
\end{equation}
where $Z^{(K)}$ denotes the final node representations.
Because of the over-smoothing problem \cite{icml2020over-smoothing}, we utilize a shallow aggregation scheme to extract the information of subgraphs in Eq.\ref{eq:hyperGCN} .
Moreover, we implement node feature encoder $f_\phi(\cdot)$ using a two-layer MLP.

\subsection{Model Training}
The classification loss $\mathcal{L}_c$ in the supervision signal is achieved through cosine similarity between $Z^{(K)}_i$ of the labeled node $v_i$ and its corresponding class prototype $\mathrm{P}_{y_i}$.
Training and inference is achieved through cosine similarities between examples and prototypes.

Ultimately, we combine classification loss $\mathcal{L}_c$ and homophilous regularizatio $\mathcal{L}_h$ into a unified training loss $\mathcal{L}$:
\begin{equation}
\label{eq:hyper-loss}
    \begin{aligned}
    \mathcal{L} =& \underbrace{
        \frac{1}{| \Omega_\mathrm{train} |}
        \sum_{i \in \Omega_\mathrm{train}} \left( 1 - \mathop{cos} \theta_{(Z^{(K)}_i, \mathrm{P}_{y_i})} \right)
    }_{\text{Classification Loss from Labeled Nodes}} 
    \\
    &+
    \lambda \; \underbrace{
        \vphantom{\sum_{j \in V_\mathrm{train}}}
        \left( 1 - \frac{1}{| \Omega_\tau |} \sum_{j \in \Omega_\tau} \frac{
                \big\Vert \, Z_j^{(K)} \big\Vert_2
            }{
                \left[ \mathcal{P}^K \mathbf{1}_{1 \times n} \right]_j
            } \right)
    }_{\text{\parbox{0.65\linewidth}{Homophilous Regularization from Unlabeled Nodes}}}
     ,
    \end{aligned}
\end{equation}
where $\lambda$ is a mixing coefficient.
The classification loss $\mathcal{L}_c$ is calculated by measuring the cosine similarity between the final representation $Z^{(K)}_i$ of the training nodes with labels (e.g., $v_i \in \Omega_\mathrm{train}$) and its corresponding prototype $\mathrm{P}_{y_i}$ .
To tackle the challenge of limited availability of high-confidence nodes during the early stages of training, we implement a warm-up strategy. This strategy entails training the model using only classification loss $\mathcal{L}_c$ during the initial phase.
After reaching a certain number of epochs, we add the homophilous regularization and utilize $\mathcal{L}$ in Eq.\ref{eq:hyper-loss} to train the model .

\section{Experiments}
We first give the experimental setup, then compare our NormProp with the state-of-the-art methods on a wide variety of semi-supervised node classification datasets in the limited labeled settings.
We analysis the impact of hyper-parameters by conducting an ablation study of NormProp.
Afterwards, we provide complexity analysis and empirical study on Ogbn-arxiv.
The code of NormProp is available at {\color{blue!50!black} \url{https://github.com/Pallaksch/NormProp} }.

\subsection{Experiment Setup}

\textbf{Datasets.}
We utilize a total of 6 real-world datasets in our experiments.
For the classical semi-supervised node classification,
we adhere to the standard fixed split \cite{icml2016planetoid-dataset} of previous work (Planetoid split, such as GCN, Violin \cite{ijcai2023violin} for Cora, Citeseer and Pubmed.
Additionally, we evaluate NormProp on few-shot semi-supervised node classification using the Cora-ML, Citeseer, MS-CS and Ogbn-arxiv with limited label, following the Meta-PN settings.
More details of the datasets can be found in the Appendix A.2.

\begin{table*}[ht]
\centering
\resizebox{1.0\textwidth}{!}{
    \begin{tabular}{l cc cc cc c}
    \toprule
    \multirow{2}{*}{\hfil \textbf{Method}}    & \multicolumn{2}{c}{Cora-ML}                           & \multicolumn{2}{c}{Citeseer}              & \multicolumn{2}{c}{MS-CS}       & \multicolumn{1}{c}{Ogbn-arxiv}             \\ 
    \cmidrule(lr){2-3} \cmidrule(lr){4-5} \cmidrule(lr){6-7} \cmidrule(lr){8-8} 
                                              & 3-shot               & 5-shot                           & 3-shot            & 5-shot              & 3-shot            & 5-shot  & 2.5\% label ratio            \\     \midrule
    MLP                                       & $41.07 \pm 0.76 $    & $51.12 \pm 0.61$    & $43.34 \pm 0.56$  & $44.90 \pm 0.60$    & $70.33 \pm 0.37$  & $79.41 \pm 0.31$     & $37.68 \pm 0.34$  \\
    GCN                                       & $48.02 \pm 0.89$     & $67.32 \pm 1.02$    & $53.60 \pm 0.86$  & $62.60 \pm 0.58$    & $69.24 \pm 0.94$  & $84.43 \pm 0.89$     & $61.36 \pm 0.51$  \\
    SGC                                       & $49.60 \pm 0.55$     & $67.24 \pm 0.86$    & $57.37 \pm 0.98$  & $61.55 \pm 0.53$    & $72.11 \pm 0.76$  & $87.51 \pm 0.27$     & $59.38 \pm 0.06$  \\
    \midrule
    LP                                        & $62.07 \pm 0.71$     & $68.01 \pm 0.62$    & $54.07 \pm 0.59$  & $55.73 \pm 1.19$    & $57.96 \pm 0.69$  & $62.98 \pm 0.61$     & $54.30 \pm 0.00$  \\
    GLP                                       & $65.57 \pm 0.26$     & $71.26 \pm 0.31$    & $65.76 \pm 0.49$  & $71.36 \pm 0.18$    & $86.10 \pm 0.21$  & $86.94 \pm 0.23$     & N/A               \\
    IGCN                                      & $66.60 \pm 0.29$     & $72.50 \pm 0.20$    & $67.47 \pm 0.29$  & $72.92 \pm 0.10$    & $85.83 \pm 0.06$  & $87.01 \pm 0.05$     & N/A               \\
    \midrule
    APPNP                                     & $72.39 \pm 0.98$     & $78.32 \pm 0.58$    & $67.55 \pm 0.77$  & $71.08 \pm 0.61$    & $86.65 \pm 0.42$  & $90.13 \pm 0.86$ & $61.60 \pm 0.28$  \\
    DAGNN                                     & $71.86 \pm 0.75$     & $77.20 \pm 0.69$    & $66.62 \pm 0.27$  & $70.55 \pm 0.12$    & $86.32 \pm 0.57$  & $90.30 \pm 0.66$     & $\underline{63.01 \pm 0.39}$  \\
    C\&S                                      & $68.93 \pm 0.86$     & $73.37 \pm 0.24$    & $63.02 \pm 0.72$  & $64.72 \pm 0.53$    & $85.86 \pm 0.45$  & $87.99 \pm 0.24$     & $56.69 \pm 0.12$ \\
    GPR-GNN                                   & $70.98 \pm 0.84$     & $75.18 \pm 0.52$    & $64.32 \pm 0.81$  & $65.28 \pm 0.52$    & $86.12 \pm 0.37$  & $90.29 \pm 0.38$     & $58.5 \pm 0.8^ \dag$     \\     \midrule
    M3S                                       & $64.66 \pm 0.31$     & $69.64 \pm 0.18$    & $65.12 \pm 0.20$  & $68.18 \pm 0.18$    & $84.96 \pm 0.18$  & $86.83 \pm 0.29$     & N/A      \\
    Meta-PN                                   & $\mathbf{74.94 \pm 0.25}$     & $ \underline{79.88 \pm 0.15}$    & $\mathbf{70.48 \pm 0.34}$  & $ \mathbf{74.14 \pm 0.50}$    & $\underline{88.99 \pm 0.29}$  & $\underline{91.31 \pm 0.22}$  & $62.66 \pm 0.14$     \\
    \midrule
    NormProp          & $\underline{73.39 \pm 0.75}$  & $ \mathbf{81.07 \pm 0.49}$          & $\underline{69.67 \pm 0.62} $ & $\underline{72.92 \pm 0.86}$   & $\mathbf{89.50 \pm 0.62}$ & $\mathbf{91.71 \pm 0.33}$ & $\mathbf{64.87 \pm 0.23}$            \\
    \bottomrule
    \end{tabular}
}
\caption{Test accuracy (\%) on few-shot semi-supervised node classification. (\lq$\dag$\rq: results are collected from published papers, \lq N/A\rq: results are not reported.)}
\label{tab:few-shot-acc}
\end{table*}

\noindent
\textbf{Baselines.}
We compare NormProp with various state-of-the-art baselines:
(\nbRoman{1}) Classical methods: GCN \cite{DBLP:conf/iclr/2017/gcn}, GIN \cite{iclr2018gin}, GraphSAGE \cite{nips2017graphsage}, SGC \cite{icml2019sgc} and GAT \cite{iclr2018gat}.
(\nbRoman{2}) Graph structure learning methods\cite{zhu2021graph-structure-learning}: GRAND \cite{nips2020grand}, GAM \cite{nips2019gam} and Violin \cite{ijcai2023violin}.
(\nbRoman{3}) Deep GNNs: APPNP \cite{iclr2019appnp}, JKNet \cite{icml2018jknet}, DAGNN \cite{kdd2020dagnn}, GCNII \cite{icml2022gcnii}, C\&S \cite{huang2020c-and-s} and GPR-GNN \cite{chien2020gpr-gnn}.
(\nbRoman{4}) Label propagation based methods: LP\cite{zhu2002label-propagation}, GLP and IGCN\cite{cvpr2019glp}. 
(\nbRoman{5}) Pseudo label based methods for few-shot node classification: M3S \cite{aaai2020M3S} and Meta-PN \cite{aaai2022meta-pn}.
The implementation details can be found in the Appendix A.4.

\noindent
\textbf{Parameter Setup.}
We implemented NormProp using Pytorch and Pytorch Geometric.
The node feature encoder $f_\phi (\cdot)$ of our proposed methods is implemented using a MLP with two layers, and the hidden size in $\{32, 64, 256, 512\}$.
We tune the following hyper-parameters: $K \in \{2, 3\}$; $\lambda \in [0, 2]$; weight decay $\in \{1e-2, 5e-3, 1e-3, 5e-4, 1e-4, 0\}$; similarity threshold $\tau \in [0, 1.0]$. We use the Adam 
\begin{table}[H]
    \centering
    \resizebox{\linewidth}{!}{
    \begin{tabular}{lccc}
        \toprule
        Method    & Cora               & Citeseer                              & Pubmed                                  \\ \midrule
        MLP       & $ 58.51 \pm 0.80 $ & $ 55.64 \pm 0.46  $                   & $ 72.71 \pm 0.61 $                      \\
        GIN       & $ 78.83 \pm 1.45 $ & $ 66.87 \pm 0.96 $                    & $ 77.83 \pm 0.42 $                      \\
        SGC       & $ 80.70 \pm 0.55 $ & $ 71.94 \pm 0.07 $                    & $ 78.82 \pm 0.04 $                      \\
        GraphSAGE & $ 81.73 \pm 0.58 $ & $ 69.86 \pm 0.62 $                    & $ 77.20 \pm 0.40 $                      \\
        GCN       & $ 82.52 \pm 0.60 $ & $ 71.02 \pm 0.83 $                    & $ 79.16 \pm 0.35 $                      \\
        GAT       & $ 82.76 \pm 0.88 $ & $ 71.87 \pm 0.53 $                    & $ 77.74 \pm 0.34 $                      \\ 
        \midrule
        JK-Net    & $ 80.35 \pm 0.58 $ & $ 67.29 \pm 1.02 $                    & $ 78.36 \pm 0.31 $                      \\
        APPNP     & $ 83.13 \pm 0.58 $ & $ 71.39 \pm 0.68 $                    & $ 80.30 \pm 0.17 $                      \\ 
        GCNII     & $ 84.17 \pm 0.40 $ & $ 72.46 \pm 0.74$                     & $ 79.85 \pm 0.34 $                      \\ 
        \midrule
        GRAND     & $ 84.50 \pm 0.30 $ & $ 74.20 \pm 0.30 $                    & $ 80.00 \pm 0.30 $                      \\
        GAM       & $ 84.80 \pm 0.06 $ & $ 72.46 \pm 0.44 $                    & $ \underline{81.00 \pm 0.09} $                      \\ 
        M3S       & $79.57 \pm 0.59$   & $69.04 \pm 0.84$  & $77.49 \pm 0.50$                     \\
        Violin    & $ \underline{84.49 \pm 0.66} $ & $ \underline{74.26 \pm 0.40} $                    & $ \mathbf{81.23} \pm \mathbf{0.42} $                      \\
        Meta-PN   & $82.07 \pm 1.34 $  & $ 71.84 \pm 1.56 $                    & $79.37 \pm 0.37 $                      \\
        \midrule
        NormProp  & $ \mathbf{85.46} \pm \mathbf{0.51} $ & $ \mathbf{74.33} \pm \mathbf{0.57} $                    & $ 80.72 \pm 1.09 $                                       \\ \bottomrule
        \end{tabular}
    }
    \caption{Test accuracy (\%) on standard semi-supervised node classification.}
    \label{tab:semi-acc}
\end{table}
\noindent
\cite{kingma2014adam} optimizer with a learning rate of $0.01$.
More details can be found in Appendix A.4.

\subsection{Experiment Results}

\noindent
\textbf{Standard semi-supervised node classification}. 
Following the typical setup of GNN experiments, we present the average and standard deviation obtained after 10 runs, as summarized in Table \ref{tab:semi-acc}.
We can observe that NormProp significantly outperform the SOTA methods in the Cora and Citeseer, even using a fixed graph convolution without adjusting graph structure.
Moreover, NormProp demonstrates a significant improvement of about $3\%$ over both GCN and SGC when applied to Cora and Citeseer, even if they utilize the same low-pass filter (e.g., $\tilde{\mathbf{D}}^{-\frac{1}{2}} \tilde{\mathbf{A}} \tilde{\mathbf{D}}^{-\frac{1}{2}}$ in Eq.\ref{eq:hyperGCN}) and undergo the same steps of propagation.
Our NormProp, in contrast to deep GNNs that enlarge the receptive field to enhance the influence of label signals, acquires additional supervisory signals via homophilous regularization solely for optimizing the model's parameters.

\noindent
\textbf{Few-shot semi-supervised node classification}. 
Following the settings in Meta-PN \cite{aaai2022meta-pn},  we report mean accuracy and $95\%$ confidence interval after 50 runs in Table \ref{tab:few-shot-acc}.
We can observe that NormProp is competitive with other SOTA methods.
In the evaluation on the large scale real-world graph ogbn-arxiv, NormProp outperforms all others under the $2.5\%$ label ratio.
Meta-PN achieves the best performance on some datasets but comes with a high complexity due to pseudo-label generation and long-distance propagation (as shown in Table \ref{tab:arxiv-time}).
On the ogbn-arxiv, our NormProp's training time is just $17\%$ of Meta-PN's, with inference time at $7\%$, further analysis is provided in Section \ref{sec:complexity}.

\begin{figure*}
    \begin{minipage}[b]{0.5\linewidth}
    \centering
    \begin{figure}[H]
    \centering
        \subfigure[Cora]{
            \centering
            \includegraphics[width=0.305\linewidth]{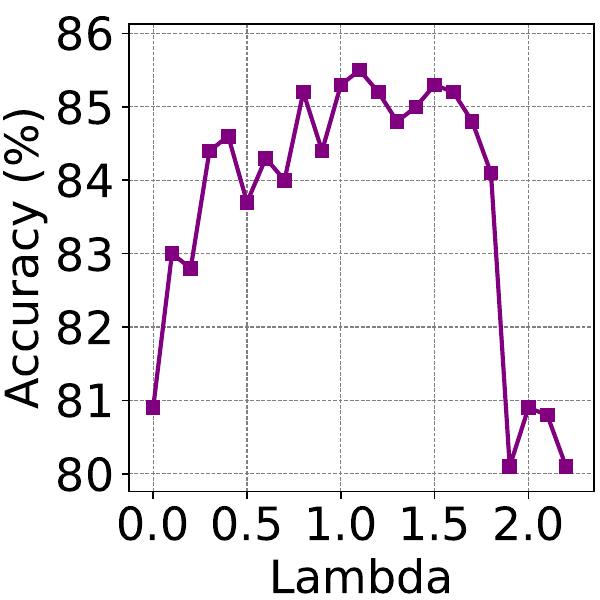}
            \label{fig:cora-mu}
        }
        \hfill
        \subfigure[Citeseer]{
            \centering
            \includegraphics[width=0.305\linewidth]{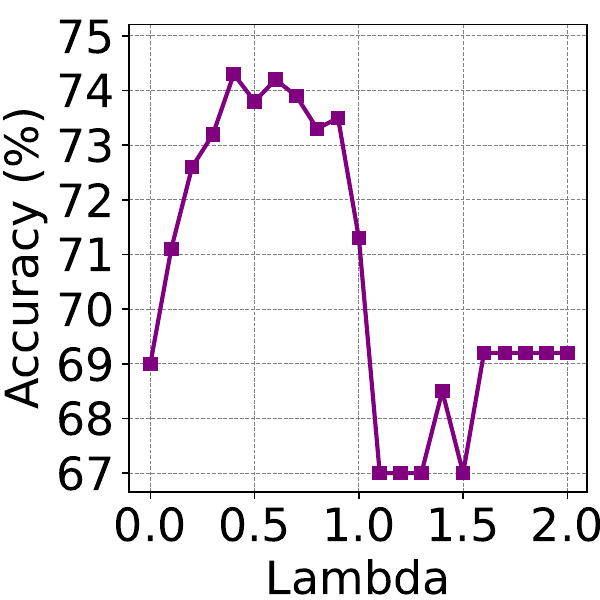}
            \label{fig:citeseer-mu}
        }
        \hfill
        \subfigure[Pubmed]{
            \centering
            \includegraphics[width=0.305\linewidth]{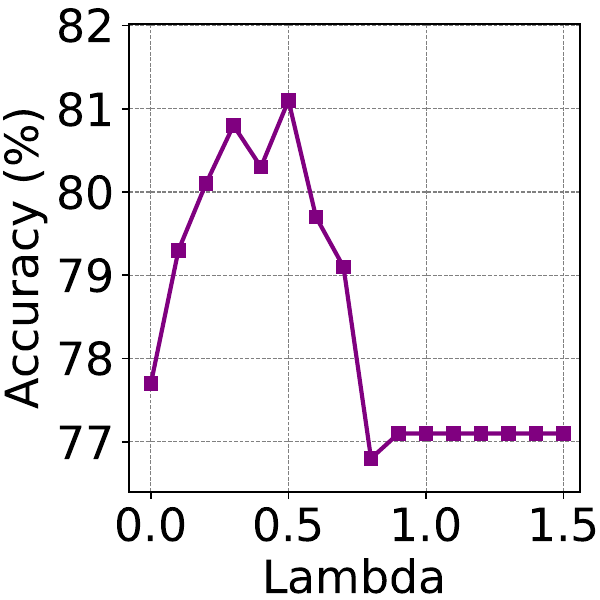}
            \label{fig:pubmed-mu}
        }
    \caption{Impact of hyper-parameter $\lambda$.}
    \label{fig:ablation-mu}
    \end{figure}
    \end{minipage}
    \hfill
    \begin{minipage}[b]{0.5\linewidth}
    \centering
        \begin{figure}[H]
        \centering
        \subfigure[Cora]{
            \centering
            \includegraphics[width=0.305\linewidth]{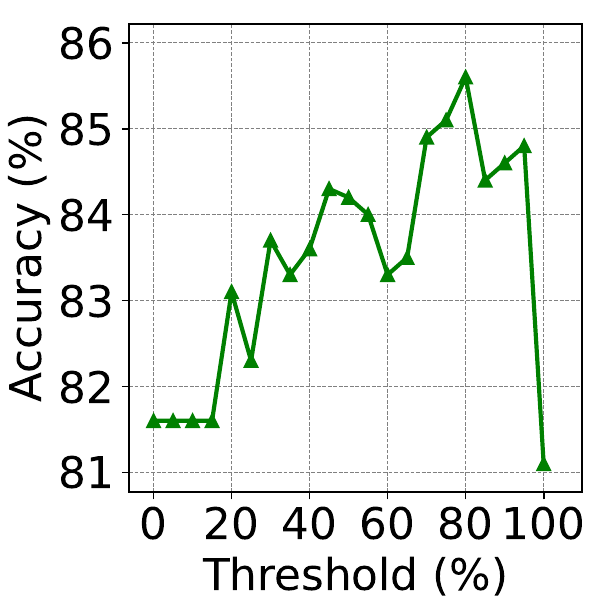}
            \label{fig:cora-threshold}
        }
        \hfill
        \subfigure[Citeseer]{
            \centering
            \includegraphics[width=0.305\linewidth]{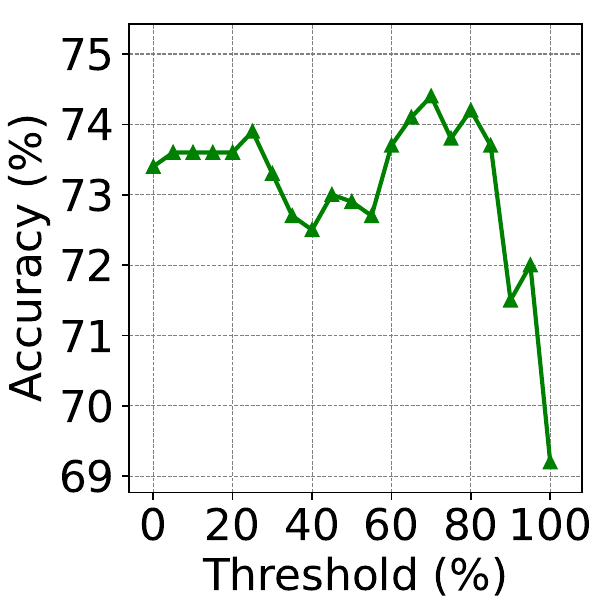}
            \label{fig:citeseer-threshold}
        }
        \hfill
        \subfigure[Pubmed]{
            \centering
            \includegraphics[width=0.305\linewidth]{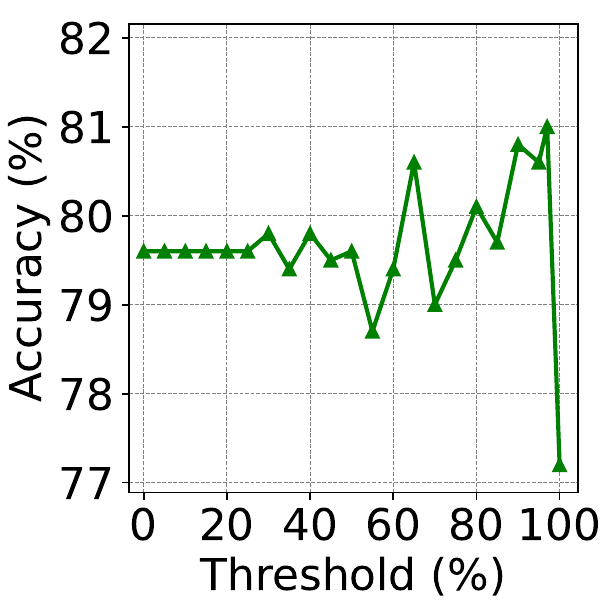}
            \label{fig:pubmed-threshold}
        }
    \caption{Impact of hyper-parameter threshold $\tau$.}
    \label{fig:ablation-threshold}
    \end{figure}
    \end{minipage}
\end{figure*}

\subsection{Ablation Study}

\begin{figure}[t]
    \centering
    \includegraphics[width=0.75\linewidth]{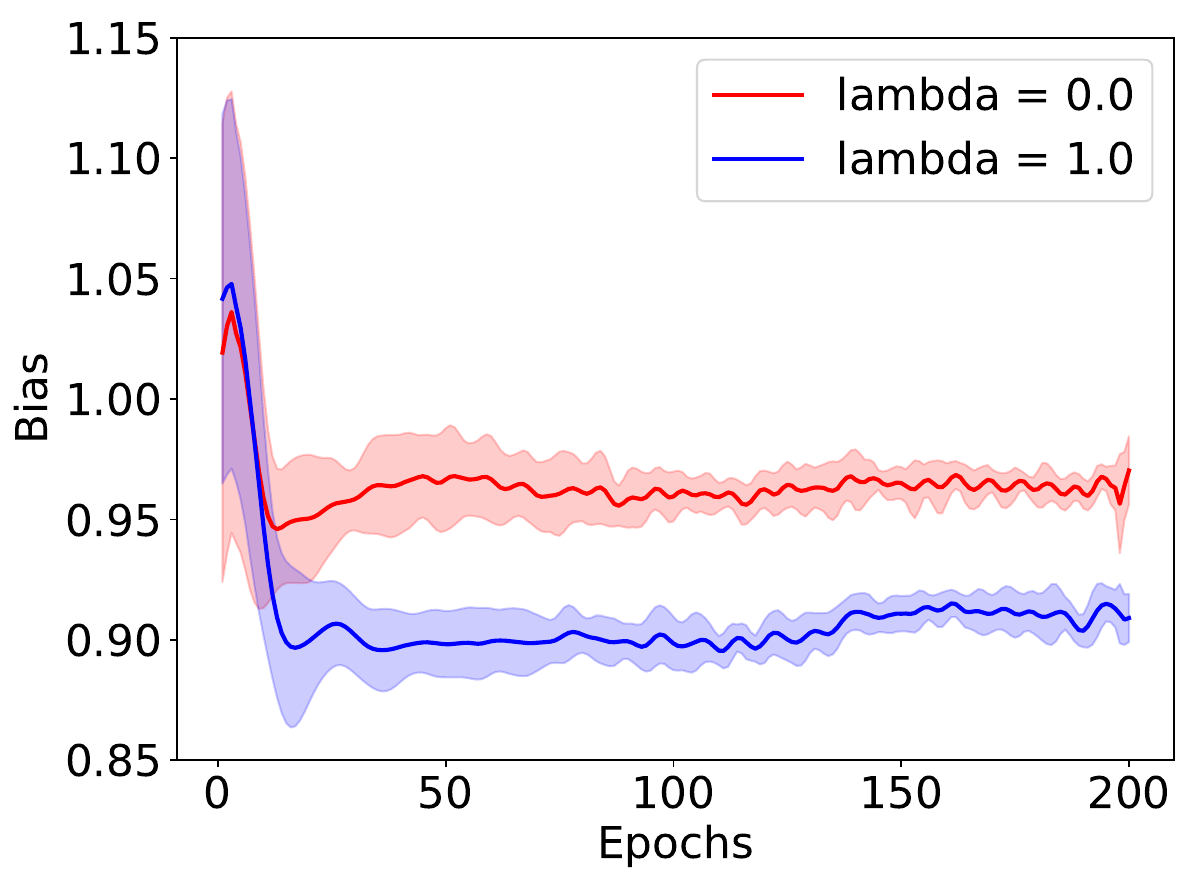}
    \captionof{figure}{Global bias of NormProp on Cora.}
    \label{fig:cora2-bias}
\end{figure}

To validate the effectiveness of $\mathcal{L}_h$, we compared the bias of NormProp for all nodes during training when $\lambda = 0$ and $\lambda = 1.0$.
We define the global bias of node representations $Z^{(k)}$ generated by NormProp: $\frac{1}{| V |} \sum_{i \in V} \left( 1 - \mathop{cos} \theta_{(Z^{(K)}_i, \mathrm{P}_{y_i})} \right)$.
Global bias is actually the calculation of classification loss $\mathcal{L}_c$ for all nodes.
The curve representing bias during training process is depicted in Figure \ref{fig:cora2-bias}.
We make the following observation: After the warm-up phase ( $>10 \textrm{ epochs}$), homophilous regularization  can effectively reduce the global bias of node representations.

To understand the hyperparameter impacts on NormProp, we conduct experiments with different values of mixing coefficient $\lambda$ and similarity threshold $\tau$.
We present the hyperparameter study in Figure \ref{fig:ablation-mu} and Figure \ref{fig:ablation-threshold} for Planetoid datasets(Cora, Citeseer, Pubmed).
When the $\lambda$ is too large, the model tends to favor trivial solutions, resulting in smoother representations of the nodes and hampers the separability between different classes.
In Figure \ref{fig:ablation-threshold}, when the $\tau$ is small, there will be more nodes with poor consistency affecting the optimization of model.
The value of $\tau$ involves the selection of unlabeled nodes involved in homophilous regularization, which is a typical quantity-quality trade-off.

\subsection{Complexity Analysis}
\label{sec:complexity}
Considering that NormProp is an end-to-end method, we analysis the time complexity of one training epoch based on Eq.\ref{eq:encode} $\sim$ Eq.\ref{eq:propagate} and Eq.\ref{eq:hyper-loss}.
We assume that the sizes of different layers in the MLP are not significantly different.
First, node representation $Z^{(0)}$ need to be prepared according Eq.\ref{eq:encode} and Eq.\ref{eq:l2-norm}. Let $|V|$ denotes the numbers of nodes, $|E|$ denotes the numbers of edges, $F'$ denotes the dim of hidden layer, $C$ denotes the class number.
The complexity of Eq.\ref{eq:encode} and Eq.\ref{eq:l2-norm} is $\mathcal{O} \left( F^2|V|\right)$.
Second, there are $K$ steps of propagation in Eq.\ref{eq:propagate}, where the time cost is $\mathcal{O}\left( K F |E| \right)$.
Third, the complexity of loss calculation Eq.\ref{eq:hyper-loss} is $\mathcal{O}\left( FC|V| \right)$.
Based on the above three components, the complexity of NormProp is $\mathcal{O}\left( F^2 |V| + KF|E| + FC|V| \right)$.
The time complexity of our method exhibits a linear relationship with both the number of nodes $|V|$ and the number of edges $|E|$.
We provide efficiency empirical study with some SOTA methods in Table \ref{tab:arxiv-time}.
NormProp demonstrates significantly greater computational efficiency compared to the other methods, both in training time and inference time. More details can be found in Appendix A.4 and A.5.

\begin{table}[t]
    \centering
    \resizebox{0.9\linewidth}{!}{
    \begin{tabular}{lcc}
        \toprule
        Method    & Training time               & Inference time                    \\
        \midrule
        GCN       & $\underline{89.97 \pm 0.18}$& $\underline{0.0598 \pm 0.0002}$   \\
        APPNP     & $107.21 \pm 0.05$           & $0.0725 \pm 0.0003$               \\
        DAGNN     & $141.05 \pm 0.02$           & $0.0942 \pm 0.0003$               \\
        Meta-PN   & $479.26 \pm 17.74$          & $0.3395 \pm 0.0001$   \\
        \midrule
        NormProp  & $\mathbf{79.45 \pm 0.04}$   & $\mathbf{0.0239 \pm 0.0002}$  \\
        \bottomrule
        \end{tabular}
    }
    \caption{Training / Inference time in seconds for different methods on ogbn-arxiv.}
    \label{tab:arxiv-time}
\end{table}

\section{Conclusion}
In this paper, we begin by analyzing the limitations of GNN generalization in the context of the semi-supervised node classification problem with limited label, focusing on the perspective of supervision signals.
In order to effectively utilize the information from unlabeled nodes to improve the generalization ability of the model, we propose a novel GNN named NormProp.
By employing ``Normalize then Propagate'' schema, the direction and Euclidean norm of node representations extracted by NormProp respectively indicate class information and consistency of aggregation.
Crucially, we provide a theoretical analysis on the upper bound of Euclidean norm.
Based on the bound of Euclidean norm and the homophily assumption, we propose homophilous regularization to improve the homophily of certain specific subgraphs.
According to our comprehensive experiments, our NormProp achieves state-of-the-art performance.
The experiments show that homophilous regularization can effectively and efficiently utilize the information from unlabeled nodes, thereby enhancing generalization ability.

\section*{Acknowledgments}
This paper is supported by the National Natural Science Foundation of China (Grant No. 62192783, 62376117), the National Social Science Fund of China (Grant No. 23BJL035), the Science and Technology Major Project of  Nanjing (comprehensive category) (Grant No. 202309007), and the Collaborative Innovation Center of Novel Software Technology and Industrialization at Nanjing University.


\bibliography{main.bib}


\end{document}